\documentclass[11pt]{article}

\usepackage[preprint]{acl}
\usepackage{wrapfig}
\usepackage{times}
\usepackage{latexsym}
\usepackage{enumitem}

\usepackage[T1]{fontenc}

\usepackage[utf8]{inputenc}

\usepackage{microtype}

\usepackage{inconsolata}

\usepackage{graphicx}

\usepackage[utf8]{inputenc} 
\usepackage[T1]{fontenc}    
\usepackage{hyperref}       
\usepackage{url}            
\usepackage{booktabs}       
\usepackage{amsfonts}       
\usepackage{nicefrac}       
\usepackage{microtype}      
\usepackage{xcolor}         
\usepackage{placeins}
\usepackage{amsmath}
\usepackage{amssymb}
\usepackage{mathtools}
\usepackage{amsthm}
\usepackage{wrapfig}
\usepackage{multirow} 
\usepackage{listings}
\lstdefinestyle{pythonstyle}{
    language=Python,
    backgroundcolor=\color{black!5},
    commentstyle=\color{green!50!black}\itshape,
    keywordstyle=\color{blue}\bfseries,
    stringstyle=\color{red!60!black},
    basicstyle=\ttfamily\footnotesize,
    showstringspaces=false,
    breaklines=true,
    frame=single,
    rulecolor=\color{black!30},
    numbers=left,
    numberstyle=\tiny\color{gray},
    xleftmargin=2em,
    framexleftmargin=1.5em
}
\usepackage{url}
\usepackage{algorithm}
\usepackage{float}
\usepackage{booktabs}
\usepackage{algorithmic}
\usepackage{wrapfig}

\newcommand{\name}[1][]{\texorpdfstring{\textsc{HeadEntropy}}{HeadEntropy}}

\title{Gradient-Stable Attention Heads Signal LLM Correctness}

\author{
  \textbf{Sophie Ostmeier\textsuperscript{1,$\dagger$}},
  \textbf{Brian Axelrod\textsuperscript{2}},
  \textbf{Maya Varma\textsuperscript{3}},
  \textbf{Asad Aali\textsuperscript{4}},
  \textbf{Yabin Zhang\textsuperscript{4}},
\\
  \textbf{Magdalini Paschali\textsuperscript{4}},
  \textbf{Sanmi Koyejo\textsuperscript{3}},
  \textbf{Curtis Langlotz\textsuperscript{4,*}},
  \textbf{Akshay Chaudhari\textsuperscript{4,*}}
\\
\\
  \textsuperscript{1}ETH AI Center, ETH Zürich, Zurich, Switzerland,
  \textsuperscript{2}Microsoft AI,
\\
  \textsuperscript{3}Department of Computer Science, Stanford University, Stanford, CA 94305, USA,
\\
  \textsuperscript{4}Department of Radiology, Stanford University, Stanford, CA 94305, USA
\\
 \small{
    \textsuperscript{$\dagger$}Work done while at Stanford University. \quad
    \textsuperscript{*}Equal contribution.
  }
\\
  \small{
    \textbf{Correspondence:} \href{mailto:sophie.ostmeier@ai.ethz.ch}{sophie.ostmeier@ai.ethz.ch}
  }
}

\begin{document}
\addtocontents{toc}{\protect\setcounter{tocdepth}{-1}}
\maketitle
\begin{abstract}
Large language models (LLMs) often generate plausible yet incorrect answers, posing risks in safety-critical settings such as medicine. Human evaluation is expensive, and LLM-as-judge approaches risk introducing hidden errors. Recent single-pass white-box methods detect contextual hallucinations using linear probes over model internals, but their generalization to out-of-domain, out-of-task, and training-free settings remains poorly understood. We introduce \name{}, a method that predicts answer correctness by measuring how susceptible each attention head's pattern is to modification under further gradient updates. The key insight is that the trace of the softmax Jacobian is a monotonic function of 2-Rényi entropy, linking attention spread to gradient stability during training, even without the reference annotation. With no training, \name{} reaches 0.736 AUROC, outperforms every training-free baseline in paired comparison and matches a hidden-state probe out-of-domain, at under 1\% of inference cost. We evaluate broadly across 5 instruction-tuned LLMs and 5 datasets spanning general knowledge, multi-hop reasoning, mathematical reasoning, claim verification, and medicine. The code is available at \url{https://github.com/StanfordMIMI/HeadEntropy}
\end{abstract}

\section{Introduction}

\begin{figure*}
    \centering
    \includegraphics[width=\linewidth]{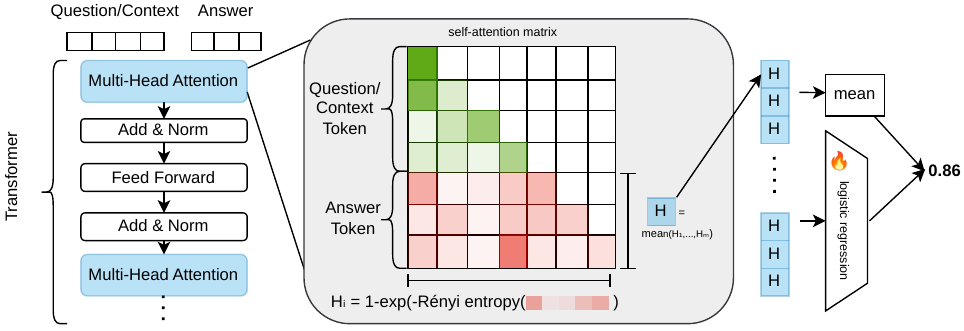}
\caption{\name{} overview. The correctness estimate is computed directly from the spread of the attention patterns, measured as the mean trace of the softmax Jacobian.}
    \label{fig:overall}
\end{figure*}

Large language models (LLMs) are increasingly being deployed in safety-critical settings, ranging from medical question answering (QA) and clinical decision support, to applications in law and scientific discovery~\citep{singhal2022large, singhal2025toward, tu2024towards, yang2024advancing, tanno2025collaboration, zambrano2025clinically}. 
However, even state-of-the-art systems can generate fluent and confident, yet \emph{incorrect} responses~\citep{wang2024prompt, sivarajkumar2024empirical, chang2025red}. 

\noindent \underline{For example:} 

\noindent \textbf{Parent:} ``\emph{I have 500\,mg aspirin tablets. Can I give one for a fever at a body weight of 24 kg?}'', 

\noindent \textbf{LLM:}``\emph{Yes, but use half a tablet. The standard antipyretic dose is 10--15\,mg/kg, so at 24\,kg roughly 250\,mg is appropriate.}'' 

\noindent The dose calculation is correct, and the model even corrects the amount the parent proposed, but the answer is not. 24\,kg indicates a child, and aspirin is avoided 
in febrile children because of its well-known link to Reye's syndrome \citep{pugliese2008reye}. 
Verifying the output with an expert may not be promptly feasible and is impractical at scale, while relying on Judge-LLMs for verification could introduce unpredictable biases or hidden errors unless the validation is also reviewed by experts~\citep{chen2024humans, thakur2024judging}.

Our work focuses on estimating whether an LLM has answered correctly using signals from \emph{inside the model itself}, computed during the same forward pass that generates the answer.
We introduce \name{}, a scalable and interpretable white-box method that uses the attention spread inside the model itself to determine correctness while generating the answer. 
In the transformer self-attention mechanism, each attention head computes attention weights by taking the inner product between query and key vectors—--a measure of their directional alignment—--and then normalizing these scores across all positions via softmax to produce a set of token weights. 
We find that certain attention patterns are stable during training, while others are unstable and tend to disappear through the course of training. We hypothesize that the presence of unstable attention patterns at inference time correlates with a generation not grounded in the training set--otherwise the training gradients would have modified the attention pattern. 

\noindent \underline{For example:} 
The LLM above reads the weight token sharply enough to compute a dose from it, since weight-based dosing is a frequent pattern used in pediatric training data and training has settled the heads that carry it. The attention that would tie the same token to the pediatric contraindication has not settled in the same way: it remains diffuse, in the regime where the attention distribution is still sensitive to perturbation and training pressure would still move it. 

\noindent Recent methods leave several questions open here. Features are selected empirically, by ablation over layers, heads, or token positions, so it is unclear when the signal should hold. Out-of-domain performance is mostly reported within one model family and the features are usually paired with a trained probe. Training-free scores such as LLM-Check and the direct ICR Score exist, but have been evaluated on few models and datasets.
\name{} addresses these questions. Its signal has a theoretically predicted direction, it requires no trained probe, and we evaluate it across tasks and model families.

\name{} makes the following contributions:
\begin{itemize}[wide=0pt, labelindent=0pt, leftmargin=*, itemsep=2pt, topsep=3pt, parsep=0pt]

\item \textbf{Attention entropy tracks gradient stability.}
The trace of the softmax Jacobian equals $1-\exp(-H_2)$, where $H_2$ is the 2-R\'enyi entropy of the attention distribution
(Appendix~\ref{app:softmax_jacobian}). Sharp (low-entropy) heads
receive small gradients and remain stable, while diffuse (high-entropy) heads stay sensitive to training pressure. We investigate this relationship empirically across pre-training checkpoints (Section \ref{sec:motivation}).

\item \textbf{A correctness predictor built on this signal.}
\name{} represents each generation by its per-head attention entropies.
High entropy at inference time indicates attention patterns that
training pressure would still change, and thus a generation that is not
grounded in the training data. This gives the feature a fixed direction:
higher entropy means a higher chance of an incorrect answer. As a
result, \name{} predicts answer correctness most consistently across dataset and model families. 

\item \textbf{The raw signal predicts correctness even without training.}
Simply averaged over heads, \name{} outperforms every competing training-free score by +8.9\% to +75.6\% relative ROC-AUC in paired comparisons across 5 models and 5 datasets (Figure~\ref{fig:paired-differences}), reaching an average of 0.74 AUROC. Out-of-domain, the untrained mean of \name{} is statistically as good as a trained hidden state probe out-of-domain ($+0.2\%$).

\end{itemize}
\section{Background}
Here, we provide a brief summary of how the Jacobian relates to Gradient Dynamics.

\subsection{The Jacobian and Gradient Dynamics}
For a vector valued function $f: \mathcal R^n \rightarrow \mathcal R^m$, the Jacobian matrix of $f(x)$ is the matrix of partial derivatives 
$$\mathbf{J}_{\mathbf{f}} = \begin{bmatrix} 
\dfrac{\partial f_1}{\partial x_1} & \cdots & \dfrac{\partial f_1}{\partial x_n} \\
\vdots & \ddots & \vdots \\ 
\dfrac{\partial f_m}{\partial x_1} & \cdots & \dfrac{\partial f_m}{\partial x_n} 
\end{bmatrix}.$$

The Jacobian can be used as a measure of how sensitive a particular set of activations or parameters are. A "large" Jacobian indicates that an identical gradient flow will change the activations greatly. We use this to identify stable versus unstable attention patterns, where we define "stable" attention patterns as those that do not change substantially with a gradient update and "unstable" attention pattern as those that change a large amount every time they appear in training. In this view, observing many unstable attention patterns may be an indicator that the current example is not like the training data.

There is no single definition of what makes a matrix such as a Jacobian "large". In this work, we use the trace of the Jacobian, noting that the softmax Jacobian is positive semidefinite (all non-negative eigenvalues), so its trace coincides with its nuclear norm (Appendix \ref{app:softmax_jacobian}).

\section{Method \label{sec:method}}

\subsection{Motivation}\label{sec:motivation}
 
\paragraph{Hypothesis.} A correct answer rests on associations that training has reinforced. We hypothesise that the attention patterns carrying such associations are ones training has settled: further gradient updates would barely move them. A pattern that continued training would still reshape is one training has not settled, and a generation resting on such patterns is less likely to be grounded in the training distribution. We want to tell the two apart at inference time.
 
Testing this directly would need the pretraining corpus: checking, for each generation, whether the associations it uses are attested in the data. That corpus is unavailable for most deployed models, and searching it once per generation would be prohibitive in any case. We need a quantity computable from the model alone.
 
\paragraph{Stability as a Jacobian trace.} Each head produces a self-attention matrix, where row $p$ applies a softmax to scores $x$. The Jacobian
\begin{equation}
  J = \frac{\partial p}{\partial x} = \mathrm{diag}(p) - pp^{\top}
\end{equation}
measures how much the attention weights change when the scores change, where the scores are the raw relevance numbers each position assigns to the others before the softmax. $J$ is positive semidefinite, so its trace sums its eigenvalues and gives a single scalar for how movable the pattern is. Large $\mathrm{tr}(J)$ means an update would readily move this head. Small means training has settled it.
 
\paragraph{The identity.} The trace has a closed form in the 2-R\'enyi entropy
$H_2(p) = -\log \sum_j p_j^2$:
\begin{equation}
  \mathrm{tr}(J) = 1 - \exp\!\left(-H_2(p)\right),
  \label{eq:identity}
\end{equation}
derived in Appendix~\ref{app:softmax_jacobian}. Thus, a head's gradient stability is fixed entirely by the spread of its attention. We can therefore read it off the forward pass without a backward pass, access to training data or additional generations.
 
\paragraph{Does training settle attention?}
If training settles the heads that carry an association, then learning an
example should lower $\mathrm{tr}(J)$ on that example, and failing to learn
it should not. We test this directly on five OLMo-2-1B pretraining
checkpoints. At each checkpoint we run one forward pass over 1000 TriviaQA
examples with their gold answers and record $\log P(\text{gold})$ together with
per-head $\mathrm{tr}(J)$. Across each of the four checkpoint transitions we
rank examples by how much $\log P(\text{gold})$ changed and call the top
decile learned, the bottom decile unlearned, and the middle 80\% the control.

\begin{figure}
\centering
\includegraphics[width=\linewidth]{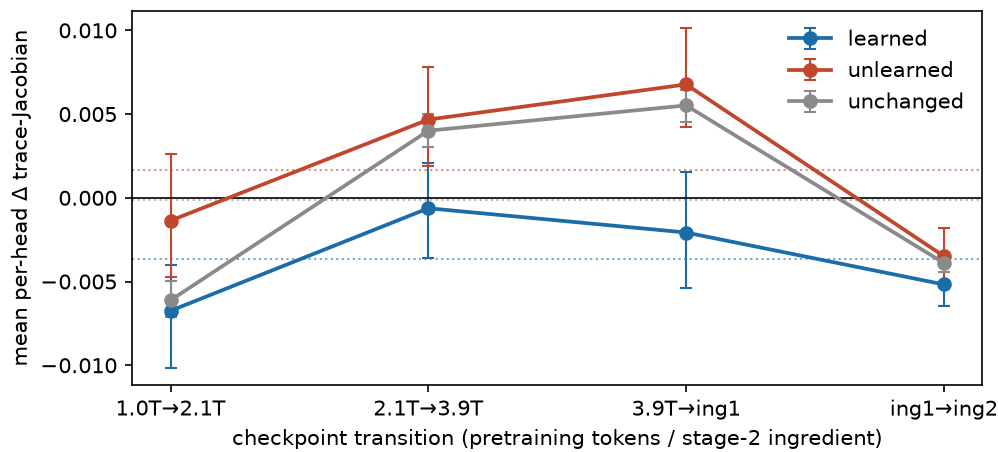}
\caption{Mean per-head change in $\mathrm{tr}(J)$ across four OLMo-2-1B checkpoint transitions, with examples grouped by the change in $\log P(\text{gold})$ over each transition. Error bars are 95\% bootstrap intervals over examples; dotted lines mark the pooled means (Wilcoxon $p = 1.8\times
10^{-12}$), while the unlearned and control groups track one another closely.}
\label{fig:olmo_transitions}
\end{figure}

Figure~\ref{fig:olmo_transitions} shows the mean change in $\mathrm{tr}(J)$
for each group. On learned examples $\mathrm{tr}(J)$ falls below the control
at every transition (pooled $-0.0035$, Wilcoxon
$p = 1.8\times10^{-12}$), whereas unlearned examples track the control, and
the more an example is learned the further its $\mathrm{tr}(J)$ falls.
Learning therefore lowers $\mathrm{tr}(J)$ during training, and \name{} reads
the same quantity at inference, where a low value indicates an answer that
rests on associations training has already settled. 
A qualitative analysis in Appendix Figure \ref{fig:correct_wrong} shows \name{} head maps for one correct and one incorrect example per dataset. For the wrong examples the heads are more yellow (higher values), especially the middle heads, indicating less settled attention patterns. 

\subsection{Computing Head Entropy}

\name{} encompasses two components: 1) computing the per-head attention entropy (Eq. \ref{eq:feature_vector}), 2a) mean over heads (Eq. \ref{eq:training_free}) or for comparison to prior methods 2b) training a logistic regression model to find predictive patterns (Eq. \ref{eq:supervised}).

Motivated by question-answering using instruction tuned LLMs, we consider an autoregressive transformer, $LLM$, with causal masking.  
The $LLM$ has $m$ attention heads in total, across all layers. We index heads by $k \in \{1,\dots,m\}$. Let $T$ be the total sequence length (i.e. question and answer tokens).

Given an input sequence $x$ (a tokenized question), the model outputs a prediction $\hat{y}$ (a tokenized answer) at nearly deterministic decoding (temperature $\approx 0$):\begin{equation} \label{eq:attention}
\hat{y} = LLM(x).
\end{equation}
 
For a given labeled question answering dataset, $\{(x_i, y_i)\}_{i=1}^N$,  we define
\begin{equation}
z_i := \mathbb{I}[\hat{y}_i = y_i] \in \{0,1\},
\label{eq:binary_label}
\end{equation}
where equality is measured with an evaluation metric (exact match and F1 score in our experiments).
Each attention head $k$ produces
\begin{equation}
A_{k} = \operatorname{softmax}(QK^\top) \in \mathbb{R}^{T \times T},
\label{eq:attention_matrix}
\end{equation}
where each row of $A_k$ is a probability distribution (i.e., rows sum to $1$).
The full tensor shape for all $m$ heads is $A \in \mathbb{R}^{m \times T \times T}$.
For each head $k$ and token $t$, we compute the $\alpha-$Rényi entropy ($\alpha=2$):
\begin{equation}
H_k^{(t)}=-\log \sum_{j=1}^T\left(A_k^{(t, j)}\right)^2
\end{equation}
and then compute the trace of the Jacobian (see Appendix \ref{app:softmax_jacobian} for the derivation).

\begin{equation}
\operatorname{trace}(J)_k^{(t)}=1-\exp\left(-H_k^{(t)}\right)
\label{eq:head_entropy}
\end{equation}

Equation \ref{eq:head_entropy} will become a feature used to predict answer correctness.

We then average over the tokens of a semantic section (e.g., question, thinking, or answer) of length $T_{\text{section}}$:
\begin{equation}
\bar{H}_{k} = \frac{1}{T_{\text{section}}} \sum_{t=1}^{T_{\text{section}}} \operatorname{trace}(J)_k^{(t)} ,
\label{eq:avg_head_entropy}
\end{equation}
where each entry, $\bar{H}_{k}$, corresponds to the \name{} of one attention head, where $k \in \{1,...,m\}$. For batched inference, we mask the rows and columns of the padding tokens. See empirical testing of aggregation techniques in the Appendix \ref{app:min_max_mean}.

This yields a fixed-dimensional feature vector for each example $i$, 
\begin{equation}
 \bar{H}_i = [\bar{H}_{1,i},...,\bar{H}_{m,i}]^\top \in  \mathbb{R}^{m}.
 \label{eq:feature_vector}
\end{equation}
In addition to improving interpretability, section-level aggregation reduces memory and compute overhead compared to token-level entropy analysis. This makes large-scale or real-time analysis more practical while preserving alignment with tasks' semantic structures.

\subsection{Training-free or supervised setting}
In the training-free setting (sometimes in literature "unsupervised"), we compute a scalar per example by taking the mean over the fixed-dimensional feature vector in Eq.\ref{eq:feature_vector},

\begin{equation}
g(\bar{H}_i) = -\frac{1}{m}\sum_{k=1}^{m} \bar{H}_{k,i} \in \mathbb{R}
\label{eq:training_free}
\end{equation}

\noindent where the negative sign orients $g$ as a score for correctness rather than against it, and is fixed a priori by the direction predicted in Section~\ref{sec:motivation}: higher head entropy indicates a higher chance of an incorrect answer.

In the supervised setting, we use fixed-dimensional feature vector for each example $i$,  $\bar{H}_i \in \mathbb{R}^m$ to train an $\ell_1$-regularized logistic regression classifier
\begin{equation}
f: \mathbb{R}^m \to [0,1]
\end{equation}
to estimate the probability of correctness:
\begin{equation}
\mathbb{P}(z_i=1 \mid \bar{H}_i) = f(\bar{H}_i) = \sigma(\beta^\top \bar{H}_i + b),
\label{eq:supervised}
\end{equation}
where $\beta \in \mathbb{R}^m$ are the coefficients, $b$ is the bias term and $\sigma$ is the logistic sigmoid. Any classifier can be used here. We chose $\ell_1$-regularized logistic regression due to simplicity and use in prior work (see more details in Section \ref{sec:ablations}). The pseudocode of the method can be found in Appendix~\ref{app:head_entropy_algorithm}.

\subsection{Computational Efficiency and Complexity}
As \name{} essentially uses summary statistics of the existing attention patterns the computational footprint is minimal compared to that of LLM inference. We document the complete complexity in Table~\ref{tab:complexity}.

\begin{table}
\centering
\caption{Computational complexity of \name{} components. \textbf{Notation:} $m$ = attention heads, $T$ = sequence length, $T_{\text{section}}$ = attention window, $N$ = training set size.}
\resizebox{\linewidth}{!}{%
\begin{tabular}{llll}
\toprule
\textbf{Component} & \textbf{Time} & \textbf{Memory} & \textbf{Notes} \\
\midrule
LLM forward pass (reference) & $O(mT^2)$ & $O(mT^2)$ & Reference \\
Feature generation (entropy) & $O(mT \cdot T_{\text{section}})$ & $O(m)$ & Negligible overhead \\
Classification (inference) & $O(m)$ & $O(m)$ & Sequence-independent \\
Training (end-to-end) & --- & $O(Nm)$ & Dominated by LLM  \\
 & &  & forward passes \\
\bottomrule
\end{tabular}
}
\label{tab:complexity}
\end{table}

Entropy features are computed using quantities already available from the softmax, adding negligible overhead to a standard forward pass. Memory remains $O(m)$ since only per-head aggregates are retained while full attention maps need not be stored. Overall, \name{} adds less than 1\% to inference cost and requires no additional forward passes beyond those typically used for generation.

\begin{table*}
    \centering
    \resizebox{\linewidth}{!}{%
    \begin{tabular}{lccccc}
        \toprule
         & TriviaQA & HotpotQA & MedMCQA & FEVER & MATH \\
        \midrule
        $\#$ Training Examples   & Sampled 30k / 130k & Sampled 30k / 98k & Sampled 30k / 183k & - & - \\
        $\#$ Validation Examples & 17,944             & 7,405             & 4,183              & 20,000             & 5,000 \\
        Task                     & Open-domain QA     & Multi-hop QA       & Multiple-choice QA & Fact verification  & Mathematical reasoning \\
        Domain                   & General           & General            & Biomedical         & General            & Mathematics \\
        Answer location          & Answer not         & Answer             & Answer in context, & Claim verified     & Answer derived \\
                                 & in context        & in context         & but not all knowledge & against evidence & via reasoning \\
        \bottomrule
    \end{tabular}%
    }
    \caption{Datasets used in experiments, with statistics on training/validation splits.}
    \label{tab:dataset}
\end{table*}
\section{Experimental Setup}

\paragraph{Models.} We evaluate three model families spanning 1.7B--32B parameters. From the reasoning-oriented Qwen3 family we use Qwen3-1.7B, Qwen3-8B, and Qwen3-32B; from the general-purpose Llama family we use Llama-3.1-8B-Instruct; and from the Gemma family we use Gemma-4-12B-it. All generations use
near-greedy decoding ($T{<}0.001$) at half-precision floating-point. We also test generalization to $T{=}1$ (Section~\ref{sec:results}, Table~\ref{tab:temperature-ablation}).

\begin{table*}
    \centering
    \resizebox{\linewidth}{!}{%
    \begin{tabular}{p{2.4cm}ccccccccccccc}
        \toprule
        & \multicolumn{4}{c}{TriviaQA} & \multicolumn{4}{c}{HotpotQA} & \multicolumn{4}{c}{MedMCQA} \\
        \cmidrule(lr){2-5}\cmidrule(lr){6-9}\cmidrule(lr){10-13}
        Model & \textbf{Acc.} & \textbf{Q} & \textbf{Think} & \textbf{Ans} & \textbf{Acc.} & \textbf{Q} & \textbf{Think} & \textbf{Ans} & \textbf{Acc.} & \textbf{Q} & \textbf{Think} & \textbf{Ans} \\
        \midrule
        Qwen3-1.7B   & 0.33 & 26 & 310 & 7  & 0.38 & 1243 & 316 & 19 & 0.29 & 47 & 519 & 6 \\
        Qwen3-8B     & 0.58 & 26 & 283 & 7  & 0.57 & 1242 & 291 & 8  & 0.39 & 47 & 505 & 6 \\
        Qwen3-32B    & 0.69 & 26 & 197 & 7  & --   & --   & --  & -- & 0.57 & 47 & 368 & 9 \\
        \midrule
        Llama-3.1-8B  & 0.45 & 25 & --  & 9  & 0.38 & 1245 & --  & 48 & 0.50 & 46 & --  & 11 \\
        \midrule
        Gemma-4-12B & 0.59 & 26 & 154 & 10 & 0.58 & 1356 & 233 & 11 & 0.43 & 48 & 501 & 12 \\
        \bottomrule
    \end{tabular}
    }
            \caption{Performance and token lengths per dataset. Accuracy (using exact match) (\textbf{Acc.}), Question (\textbf{Q}), Think chunk (\textbf{Think}), and Answer (\textbf{Ans}) token length. }
                \label{tab:perf_tokens_by_dataset}
\end{table*}

\paragraph{Datasets.} We use five benchmarks spanning open-domain QA (TriviaQA, \citealp{joshi2017triviaqa}), multi-hop QA with long context (HotpotQA distractor setting, \citealp{yang2018hotpotqa}), domain-specific
multiple-choice QA (MedMCQA, \citealp{pmlr-v174-pal22a}), fact verification (FEVER, \citealp{thorne2018fever}), and mathematical reasoning (MATH, \citealp{hendrycks2021measuring}). For TriviaQA, HotpotQA, and MedMCQA we sample uniformly at random with a fixed seed 30k training examples with a max of 768 generated tokens. Truncated answers were excluded from the training set only. We use the given validation split for evaluation. FEVER and MATH given validation split are held out and used for out-of-distribution evaluation (Table~\ref{tab:dataset}). The prompts can be found in  Appendix~\ref{app:prompt_hot}, \ref{app:prompt_med}, \ref{app:prompt_triv}).
We define binary correctness via the standard metric for each dataset: exact match for TriviaQA, HotpotQA, and MATH; label match for FEVER; option match for MedMCQA (Eq. \ref{eq:binary_label}). The overall accuracy of the models on the QA task is given in Table \ref{tab:perf_tokens_by_dataset}.

\paragraph{Baselines.} We compare against nine single-forward-pass methods,
covering both raw scalar statistics and learned linear probes:
(i)~token entropy \citep{kadavath2022language};
(ii)~attention score, the kernel-similarity statistic from
LLM-Check \citep{sriramanan2024llm} (picked layer: 0.8 * total number of layers);
(iii)~mean hidden states, the average final-layer hidden
state used as a raw signal;
(iv)~mean Lookback Lens ratio, the average
context-vs-generation attention ratio \citep{chuang-etal-2024-lookback};
(v)~verbalized Uncertainty (see Appendix \ref{app:prompt_triv},\ref{app:prompt_med}, \ref{app:prompt_hot});
(vi)~mean ICR score, the information-flow statistic of
\citet{zhang2025icr};
(vii)~Hidden State LR, an $\ell_1$-regularized logistic
regression on final-layer hidden states, following
\citet{xie2024calibrating};
(viii)~Lookback Lens LR, the same probe trained on
per-head Lookback Lens features as in \citep{chuang-etal-2024-lookback};
(ix)~ICR LR, the same probe trained on the per-layer ICR scores.
Baselines (i)--(vi) are training-free; (vii)--(ix) use the same logistic regression probe setup as \name{}. Multi-pass methods such as Semantic Uncertainty \citep{kuhn2023semantic} and INSIDE \citep{cheninside} are not directly comparable since they require additional generations or auxiliary models; we report their published TriviaQA AUROC for context in the Appendix Table~\ref{tab:triviaqa-comparison}.

\paragraph{Metrics.} We measure how well each score separates correct from incorrect model answers with AUROC. Scores are reported with their intended orientation, if known from literature, i.e. a higher ratio for Lookback lens indicates an answer being more grounded in the context. 
\begin{figure*}[t]
  \centering
  \includegraphics[width=\linewidth]{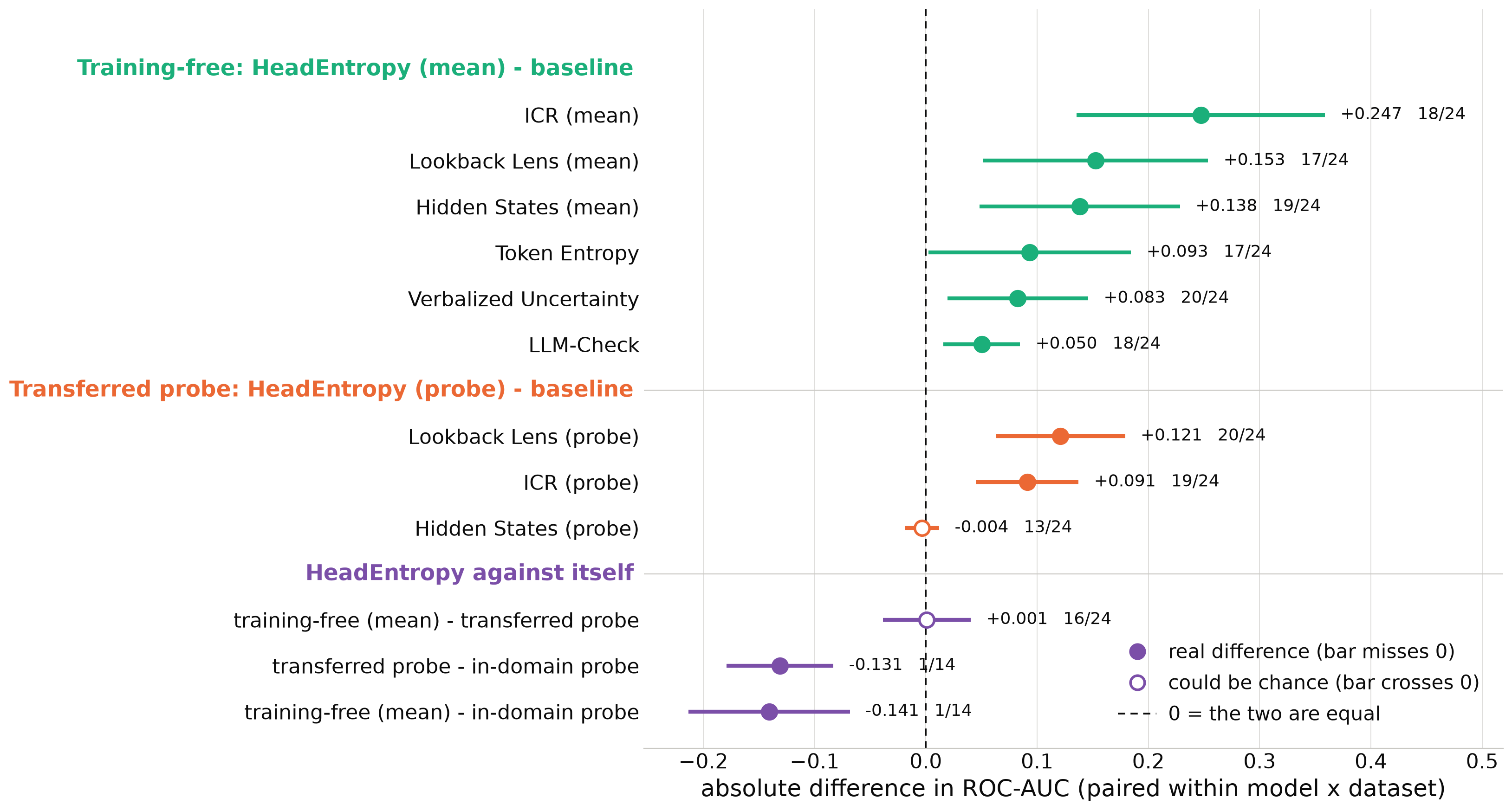}
  \caption{Paired differences in ROC-AUC. Each row is a difference taken within
    every (model, dataset) cell and then averaged, so model and dataset difficulty
    cancel instead of inflating the spread. The bar is the 95\% interval of those
    per-cell differences, taken from the Student's $t$ quantile at $n-1$ degrees of
    freedom; For
    Hidden States, ICR and Lookback Lens in the training-free setting (only there),
    $1-\mathrm{AUC}$ was applied after seeing the results; the orientation of \name{} is fixed a priori by Eq.~\ref{eq:training_free}.
  }
  \label{fig:paired-differences}
\end{figure*}

\section{Results}
\label{sec:results}

With no training at all, the mean of \name{} over heads reaches 0.736 AUROC
across 5 models and 5 datasets, beats every training-free baseline in paired
comparison (Figure~\ref{fig:paired-differences}), and matches a hidden-state probe transferred across domains (0.739). A probe helps in-domain and adds nothing under domain shift.

We then ask what supervision adds to this raw signal, reporting results as a
ladder of increasing supervision:
(i)~the raw signal with no probe (Section~\ref{sec:results-unsup});
(ii)~what a trained probe adds over that raw signal, in-domain and under
domain shift (Section~\ref{sec:results-probe});
(iii)~where a probe is available, how \name{} compares against probes on
other features (Section~\ref{sec:results-id}).

\subsection{Training-free correctness prediction}
\label{sec:results-unsup}

A training-free signal requires no labeled data, no per-domain retraining, and no
fitting procedure that could overfit to a particular distribution, properties
especially relevant in safety-critical deployments. We compare the mean of per-head \name{} values (Eq.~\ref{eq:training_free})
against six training-free scores: token entropy, attention score (LLM-Check),
verbalized uncertainty, mean hidden states, mean Lookback Lens ratio, and the mean ICR
score (Table~\ref{tab:ood_results_dataset}).

\name{} reaches the highest average AUROC, 0.736, ahead of attention score (0.685),
verbalized uncertainty (0.655), token entropy (0.646), mean hidden states (0.597),
mean Lookback Lens (0.586) and mean ICR (0.491). In the  paired comparison (Figure \ref{fig:paired-differences})
\name{} is significantly better compared to the baselines, 
from $+0.05$ AUC over attention score ($+8.9\%$ relative) to $+0.25$ over mean ICR
($+75.6\%$), and it wins in 17 to 20 of the 24 pairs (Figure~\ref{fig:paired-differences},
top block).

Across domains, \name{} ranks first or second in every case. It leads on MedMCQA and HotpotQA, is level with the attention score on MATH, and is second behind Lookback Lens on TriviaQA and FEVER. The other methods rank lower in at least one domain. FEVER is the hardest domain for all training-free scores (Table~\ref{tab:ood_results_dataset}).

The advantage varies across backbones. \name{} is the strongest training-free score on the three Qwen3 models, while mean ICR leads on Gemma-4-12B and token entropy on Llama-3.1-8B. Overall, \name{} has the highest average and varies least across backbones (Appendix Table~\ref{tab:ood_results}).


\subsection{What the probe buys}
\label{sec:results-probe}

Comparing \name{} against itself across the three regimes (supervised in-domain, supervised out-of-domain, training-free) shows the contribution of
the trained probe (Figure~\ref{fig:paired-differences}, bottom block). For completeness, Appendix Figure~\ref{fig:auc-by-method-regime} gives the absolute levels these differences are taken from, for every method and regime.

Under domain shift the \name{} probe contributes nothing measurable. The training-free mean and the probe transferred from another dataset differ by $+0.001$ AUC ($+0.2\%$ relative). 
Once the probe is applied outside the domain it was fitted on, it recovers no more of the signal than plain averaging does. 
Fitting in-domain does buy $0.141$ AUC and almost all of that gap is the domain shift itself rather than the fit, since the probe out-of-domain gives up $0.131$ AUROC compared to the in-domain one, which is less than the $0.176$ for Lookback Lens, $0.274$ for ICR and $0.299$ for hidden states. 
Hidden states match \name{} when a probe is fitted (Section~\ref{sec:results-id}) and lose more than twice as much when it is taken away. 
This suggests that higher entropy is tied a priori to a less settled attention pattern, the direction of the feature is known before any data is seen, and the probe is left with little to add beyond that direction when tested out-of-domain.
\begin{table*}
    \centering
    \resizebox{\linewidth}{!}{%
    \begin{tabular}{lcccccc}
        \toprule
        Method & TriviaQA & HotpotQA & MedMCQA & FEVER & MATH & Average \\
        \midrule
        Token Entropy          & 0.609 & \underline{0.657} & 0.684 & 0.578 & 0.700 & 0.646 \\
        Verbalized Uncertainty & 0.657 & 0.621 & 0.689 & 0.619 & 0.688 & 0.655 \\
        Attention Score        & 0.659 & 0.616 & \underline{0.721} & 0.549 & \bfseries  0.882 & \underline{0.685} \\
        Hidden States (mean)   & 0.579 & 0.516 & 0.601 & 0.575 & 0.712 & 0.597 \\
        Lookback Lens (mean)   & \bfseries 0.719 & 0.596 & 0.596 & \bfseries  0.645 & 0.376 & 0.586 \\
        ICR (mean)             & 0.482 & 0.491 & 0.500 & 0.509 & 0.474 & 0.491 \\
        \midrule
        \name{} (mean)         & \underline{0.693} & \bfseries  0.668 & \bfseries  0.806 & \underline{0.632} & \underline{0.879} & \textbf{0.736} \\
        \bottomrule
    \end{tabular}
    }
    \caption{Training-free performance (AUROC) across evaluation datasets, averaged over
    the five models}
    \label{tab:ood_results_dataset}
\end{table*}

\subsection{Supervised correctness prediction}
\label{sec:results-id}

\paragraph{In-domain.}
Trained and evaluated on the same dataset, a \name{} probe averages 0.867 AUROC across models and datasets, statistically indistinguishable from Hidden States LR (0.868; paired difference $-0.000$) and ahead of Lookback Lens LR ($0.818$; $+0.051$) and the ICR probe (0.767; $+0.102$); see Appendix~\ref{app:supervised_breakdown},
Figure~\ref{fig:auc-by-method-regime}.

\paragraph{Comparison with prior work.}
Appendix Table~\ref{tab:triviaqa-comparison} places \name{}'s TriviaQA AUROC on Llama-3.1-8B (0.879) next to numbers reported by prior hallucination-detection and uncertainty-estimation methods. Cross-paper comparison is loose, since setups, backbones and evaluation protocols differ. \name{} is in the range of reported single-pass numbers (LapEigvals 0.889, Kadavath et al.\ 0.864) and above the multi-pass Semantic Uncertainty (0.828) and Semantic Energy (0.748).

\subsection{Ablations}
\label{sec:ablations}
Appendix ablations show that performance is stable with 20\% of the training data, at regularization strength $c$=0.1, and under sampling at temperature even with $T$=1 (AUROC shifts of at most 0.05 on TriviaQA and HotpotQA; Appendices~\ref{app:c_ablation}, \ref{app:data_ablation}, Table~\ref{tab:temperature-ablation}). The exception is MedMCQA, where AUROC drops by 0.20 to 0.24. However, model accuracy there falls below chance (0.18 and 0.12 vs. 0.25), so the generations themselves are unreliable rather than the signal.
We justify mean aggregation over min/max (Appendix~\ref{app:min_max_mean}). 
Regarding the probe architecture, a logistic regression model performed on par with XGBoost and MLP probes, with at most a negligible difference (Appendix \ref{app:xgboost_mlp}).
We show, via SHAP analysis and a layer ablation, that middle layers carry the most signal (Appendices~\ref{app:computation_shaply}, \ref{app:layer_ablation}).

\section{Related Work}
Two fundamental goals motivate this study: (i) \emph{uncertainty estimation} and (ii) \emph{hallucination detection}. 
Uncertainty estimation provides a measure of how much we can trust a model’s output. 
Hallucination detection, on the contrary, aims to identify statements in generated text that are unsupported or contradict established evidence. 
Our work translates broader uncertainty estimation and hallucination detection into the more concrete objective of estimating answer-level correctness, that is, the probability that an answer produced by an LLM given an input is correct under a task metric (e.g., exact match, F1 score) (Eq. \ref{eq:binary_label}).

Prior work differs in how much of the LLM’s internal state is used to fulfill these goals: (i) black-box methods rely only on outputs, often requiring multiple passes or auxiliary models; (ii) gray-box methods access shallow signals such as final-layer logits; and (iii) white-box methods leverage deeper representations, such as hidden states or attention maps. 

\textbf{Black-box methods.} 
These methods rely solely on the LLM's input and outputs and often require multiple forward passes or auxiliary models, making them expensive and sensitive to domain shift. SelfCheckGPT~\citep{manakul2023selfcheckgpt} detects uncertainty via disagreement across sampled completions, while Chain-of-Verification~\citep{dhuliawala2024chain} probes consistency through follow-up questions. Other approaches, such as semantic uncertainty~\citep{kuhn2023semantic}, use embedding models to group paraphrases, introducing the reliance on well-aligned external tools. LLM-as-a-Judge methods~\citep{chen2024humans, thakur2024judging} delegate evaluation to separate LLMs, potentially introducing biases and weak calibration guarantees. 

\textbf{Gray-box methods.} Gray-box methods occupy a middle ground, drawing on limited internal signals such as token log probabilities or final hidden states. By doing so, they enable single-pass confidence estimation. For example, predictors trained over logits can temper the well-documented tendency of LLMs toward overconfidence \citep{groot2024overconfidence}, while filters based on final-layer statistics can anticipate distributional shifts that undermine reliability \citep{pouget2025suitability}. Likewise, conformal prediction methods extend log-probability signals into calibrated sets of plausible completions \citep{angelopoulos2021gentle}, offering a principled notion of coverage. Despite these strengths, gray-box approaches remain constrained by their reliance on final-layer information, which often calibrates poorly across domains and reduces interpretability to heuristics like “low log-probability” \citep{geng2024surveyconfidenceestimationcalibration}.

\textbf{White-box methods.} These methods leverage richer internal signals such as layer-wise hidden states, attention maps, or neuron activations, often improving accuracy while introducing additional design choices. Several works~\citep{azaria2023internalstatellmknows, kadavath2022language} have demonstrated that an LLM's hidden states encapsulate latent knowledge about hallucinations. \citet{cheninside} sample multiple responses and use the log-determinant of their embedding covariance as a correctness signal. \citet{snyder2024early} employ gated recurrent units (GRUs) and multilayer perceptrons (MLPs) over internal representations to predict correctness. For more details on this line of work, please see Appendix~\ref{app:hidden_state_related_work}. 
LLM-Check~\citep{sriramanan2024llm} combines multiple internal signals, including attention statistics, to construct a correctness metric. Distinct from the works mentioned above, a small group of methods focuses only on the attention heads of a single sample.
Lookback lens detects contextual hallucinations by measuring the location of the attention mass by the ratio of attention weights on the context versus newly generated tokens ~\citep{chuang-etal-2024-lookback}.
While AggTruth focuses on contextual hallucination detection in RAG settings, analyzing how attention is distributed across passage tokens using multiple aggregation techniques ~\citep{matys2025aggtruth}.
\name{} targets final-answer correctness rather than contextual 
hallucination, and uses the spread of the full attention distribution rather than the location of attention mass.
\section{Conclusion}
\name{} predicts answer correctness with no training, no calibration, and a direction fixed before any data is seen. Its untrained mean reaches 0.736 AUROC, the level of a hidden-state probe transferred across domains, at under 1\% of inference cost. On thinking models the signal is strongest (0.80 to 0.82 AUROC); on the non-thinking Llama-3.1-8B it weakens, which we leave as an open question.
\section*{Limitations}
The connection between attention entropy and gradient dynamics is a
motivation for the feature. $\mathrm{tr}(J)$ says how
far a pattern moves under a perturbation of its own scores; the size of a real update also depends on the gradient reaching that head, which we do not measure.
\name{} supports any binary correctness label (including GPT-as-judge), but we restrict to deterministic metrics for reproducibility. Open-ended tasks like summarization and dialogue lack this binary notion and represent a complementary direction, as does code generation where automated verification is available \citep{chuang-etal-2024-lookback, matys2025aggtruth}.
\name{} uses the full context (average 26 tokens on TriviaQA to 1242 on HotpotQA). A sliding-window variant in the style of Lookback Lens is a natural extension but not tested here.


\FloatBarrier

\bibliography{custom}

\appendix
\clearpage
\onecolumn
\raggedbottom
\addtocontents{toc}{\protect\setcounter{tocdepth}{2}}
{\renewcommand{\contentsname}{Appendix Contents}\tableofcontents}

\section{Use of LLM}
We use LLMs to check grammar and typos. 
\section{Prompts}
\subsection{TriviaQA Prompt}
\label{app:prompt_triv}
\begin{lstlisting}[style=pythonstyle]
def build_triviaqa_prompt(item):
    question = item["question"]
    
    system_message = (
        "You are a trivia expert. Please answer questions in exactly this format:\n"
        "Answer: [1-3 words only]\n"
        "Certainty: [0-100]\n\n"
    )
    
    user_prompt = f"Question: {question}"
    
    return system_message, user_prompt
\end{lstlisting}

\subsection{MedMCQA Prompt}
\label{app:prompt_med}
\begin{lstlisting}[style=pythonstyle]
def build_ind_prompt(item):
    
    system_message = (
        "You are a medical expert. Please answer questions in exactly this format:\n"
        "Answer: [repeat correct option]\n"
        "Certainty: [0-100]\n\n"
    )
    user_prompt = (
        "Question:"
        + item["question"].split(". ")[-1]
        + "\n"
        + "Options:"
        + " ".join(str(item[op]) + "\n" 
                   for i, op in enumerate(["opa", "opb", "opc", "opd"]))
    )
    return system_message, user_prompt
\end{lstlisting}

\subsection{HotpotQA Prompt}
\label{app:prompt_hot}
\begin{lstlisting}[style=pythonstyle]
def build_hotpot_prompt(item):
    system_message = (
        "You are a helpful assistant. "
        "Answer the question using the information in the provided passages."
        "Please answer questions in exactly this format:\n"
        "Answer: [1-5 words only]\n"
        "Certainty: [0-100]\n\n"
    )
    list_sentences = []
    for sentence_list in item["context"]["sentences"]:
        list_sentences.extend(sentence_list)

    context_text = "\n".join(sentence for sentence in list_sentences)

    user_prompt = (
        "Context: "
        f"{context_text}\n\n"
        f"Question: {item['question']}\n"
    )

    return system_message, user_prompt
\end{lstlisting}

\section{Model dimensions}

\begin{table}[H]
\caption{Feature dimensionality per model. $d$: hidden size (dimension of a mean-pooled hidden-state feature); $h$: attention heads per layer; $L$: layers; $m = hL$: total heads, the dimension of the \name{} feature vector. The Qwen3 and Llama models use $d_{\text{head}}=128$ and Gemma-4-12B uses $d_{\text{head}}=256$; Qwen3-32B and Gemma-4-12B are the two where $h \cdot d_{\text{head}} \neq d$. Values from each model's \texttt{config.json} on Huggingface.}
\begin{tabular}{lrrrrr}
\toprule
Model & $d$ & $h$ & $L$ & $m$ & $d/m$ \\
\midrule
Qwen3-1.7B    & 2048 & 16 & 28 & 448  & 4.6 \\
Llama-3.1-8B  & 4096 & 32 & 32 & 1024 & 4.0 \\
Qwen3-8B      & 4096 & 32 & 36 & 1152 & 3.6 \\
Qwen3-32B     & 5120 & 64 & 64 & 4096 & 1.3 \\
Gemma-4-12B   & 3840 & 16 & 48 & 768  & 5.0 \\
\bottomrule
\end{tabular}
\label{tab:dims}
\end{table}
\FloatBarrier
\section{Related work on hidden states as features for hallucination detection}
\label{app:hidden_state_related_work}

A growing line of work uses hidden states as features for predicting 'hallucinations', a term used loosely across this literature to cover varied failure modes, including factual errors, inconsistencies, and errors flagged by LLM judges. In contrast, \name{} uses only the attention distribution: it is lower-dimensional, interpretable, automated across layers, and requires a single forward pass at inference time.

\citet{kossen2024semantic}'s SEPs trains hidden state probes to predict semantic entropy, while \name{} uses attention entropy to predict answer correctness directly under standard QA metrics. \name{} offers three advantages over a SEPs-style hidden state probe for the given task: First, we provide a formal motivation connecting 2-Rényi attention entropy to training gradient dynamics via the trace (equivalently, nuclear norm) of the softmax Jacobian. SEPs demonstrates empirically that hidden states encode semantic uncertainty, while \name{} explains why attention entropy relates to gradient stability. Second, \name{} avoids manual layer and token position selection entirely. Attention entropy is available at every layer and head, and the sparse classifier automatically identifies informative heads, whereas SEPs require ablations to find optimal layer and token positions.
Third, the feature vector has dimensionality m (number of attention heads, typically for open source models 16-64) versus hidden state dimensionality d (typically 1024-8192). Supervising \name{} on binary semantic entropy labels (analogous to Kossen et al.) is an interesting future direction. For the scope of this work we focus on exact match and F1 as correctness metrics.

\citet{cencerrado2025no} focuses on predicting answer correctness before generation using linear probes on hidden states with a heuristic motivation. Although our work emphasizes the attention distribution using a formal motivation, the task is comparable. Similarities: the best-performing assessor (trained on embeddings from GPT) is logistic regression, signal increases with larger models, and middle layers perform slightly better. Key differences: their core claims focus on OOD generalization for predicting answer correctness from question-only hidden states, a distinct goal from ours. Results are reported on a subsampled training set, making direct comparison non-trivial as open-source models are almost certainly trained on it. They use careful per model layer selection, whereas \name{} automates this process. 

\citet{ji2024llm} addresses a similar task; however, it focuses on self-awareness of uncertainty by examining how internal states indicate whether the model has seen the query during training, a question related to our paper. We focus on a formal motivation through the lens of training stability and connect feature generation directly to training dynamics, with their hidden-state approach serving as one of our baselines. They employ a more complex multilayer perceptron (MLP) variant. 

\citet{huang2025internal} investigates whether internal causal mechanisms can robustly predict model correctness in out-of-distribution settings, covering tasks such as symbol manipulation, knowledge retrieval, and instruction following. While related to our goal, we examine more closely the contribution of each attention head using a logistic regression model and note that generalization across tasks exists, but with some task-dependent component. 

\citet{beigi2024internalinspector} is in the same field as the studies mentioned above. One focus here is predictor model selection and contrastive loss construction to cope with high-dimensional inputs from hidden, activation, and attention states across all layers. 

Unlike methods analyzing static, high-dimensional hidden states, 
\citet{zhang2025icr} focuses on the dynamic update process of the residual stream, which proves more effective at identifying when a model deviates from factual accuracy across layers. The ICR Probe outperforms prior hidden-state baselines and provides improved interpretability for understanding hallucination mechanisms.

\begin{table}[H]
\centering
\caption{TriviaQA in-distribution correctness-prediction AUROC across hallucination detection methods. All numbers except \name{} are reported from the original papers and in-distribution; setups and models differ across rows and numbers are not strictly comparable. $^\dagger$Reimplemented result; the original paper does not report TriviaQA.}
\label{tab:triviaqa-comparison}
{%
\begin{tabular}{llcccc}
\toprule
Method & Model & Train & Test & Training & TriviaQA \\
 & & Gen. & Gen. & Free & AUROC \\
\midrule
Semantic Uncertainty \citep{kuhn2023semantic}    & OPT-30B       & ---  & $>$1 & \checkmark & 0.828 \\
Semantic Energy \citep{ma2025semantic}           & Qwen3-8B      & ---  & $>$1 & \checkmark & 0.748 \\
\addlinespace
Kadavath et al.\ \citep{kadavath2022language}    & 52B (P(IK))   & $>$1 & 1    & $\times$   & 0.864 \\
SEP \citep{kossen2024semantic}                   & Llama-2-7B    & $>$1 & 1    & $\times$   & $<$0.85 \\
\addlinespace
Snyder et al.\ \citep{snyder2024early}           & OpenLlaMA-7B  & 1    & 1    & $\times$   & 0.71 \\
LapEigvals \citep{binkowski2025hallucination}    & Llama-3.1-8B  & 1    & 1    & $\times$   & 0.889 \\ \midrule
ICR Probe \citep{zhang2025icr}                   & Llama-3       & 1    & 1    & $\times$   & 0.7325 $^\dagger$ \\
\addlinespace
LLM-Check \citep{sriramanan2024llm}         & Llama-2-7B    & ---  & 1    & \checkmark & 0.742$^\dagger$ \\

LookbackLens                                     & Llama-3.1-8B  & 1    & 1    & $\times$   & 0.883 $^\dagger$ \\
Hidden State LR                                  & Llama-3.1-8B  & 1    & 1    & $\times$   & 0.858 $^\dagger$ \\
\name{}                                          & Llama-3.1-8B  & \textbf{1} & \textbf{1} & $\times$   & 0.879 \\
\bottomrule
\end{tabular}
}
\end{table}

\section{Training-free performance per model}
\label{app:unsup_per_model}
Table~\ref{tab:ood_results} breaks the training-free comparison of
Section~\ref{sec:results-unsup} down by backbone, averaging AUROC over the five
evaluation datasets. \name{} has the highest average (0.741) and is the strongest
training-free score on all three Qwen3 models, by 0.05 to 0.11 AUROC over the
runner-up. On Gemma-4-12B mean ICR (0.773) and attention score (0.740) are
slightly ahead of \name{} (0.733). On Llama-3.1-8B every attention- or
hidden-state-based score is close to chance and token entropy (0.723) is the only
informative signal; this non-thinking model is the hardest backbone for all methods.
\begin{table}[H]
    \centering
    \resizebox{\linewidth}{!}{%
    \begin{tabular}{lcccccc}
        \toprule
        Method & Llama-3.1-8B & Gemma-4-12B & Qwen3-1.7B & Qwen3-32B & Qwen3-8B & Average \\
        \midrule
        Token Entropy         & \textbf{0.723} & 0.706 & 0.580 & 0.576 & 0.628 & 0.642 \\
        Verbalized Uncertainty & \underline{0.543} & 0.645 & 0.708 & 0.613 & \underline{0.762} & 0.654 \\
        Attention Score       & 0.501 & \underline{0.740} & \underline{0.769} & \underline{0.691} & 0.743 & \underline{0.688} \\
        Hidden States (mean)  & 0.517 & 0.730 & 0.346 & 0.671 & 0.750 & 0.603 \\
        Lookback Lens (mean)  & 0.516 & 0.499 & 0.656 & 0.611 & 0.652 & 0.587 \\
        ICR (mean)            & 0.522 & \textbf{0.773} & 0.271 & 0.465 & 0.421 & 0.490 \\
        \midrule
        \name{} (mean)        & 0.536 & 0.733 & \textbf{0.815} & \textbf{0.800} & \textbf{0.821} & \textbf{0.741} \\
        \bottomrule
    \end{tabular}
    }
    \caption{Training-free performance (AUROC) across models, averaged over the five evaluation datasets. Best per column in bold, second best underlined.}
    \label{tab:ood_results}
\end{table}

\section{Supervised performance}
\label{app:supervised_breakdown}
\begin{figure}[H]
    \centering
    \includegraphics[width=\linewidth]{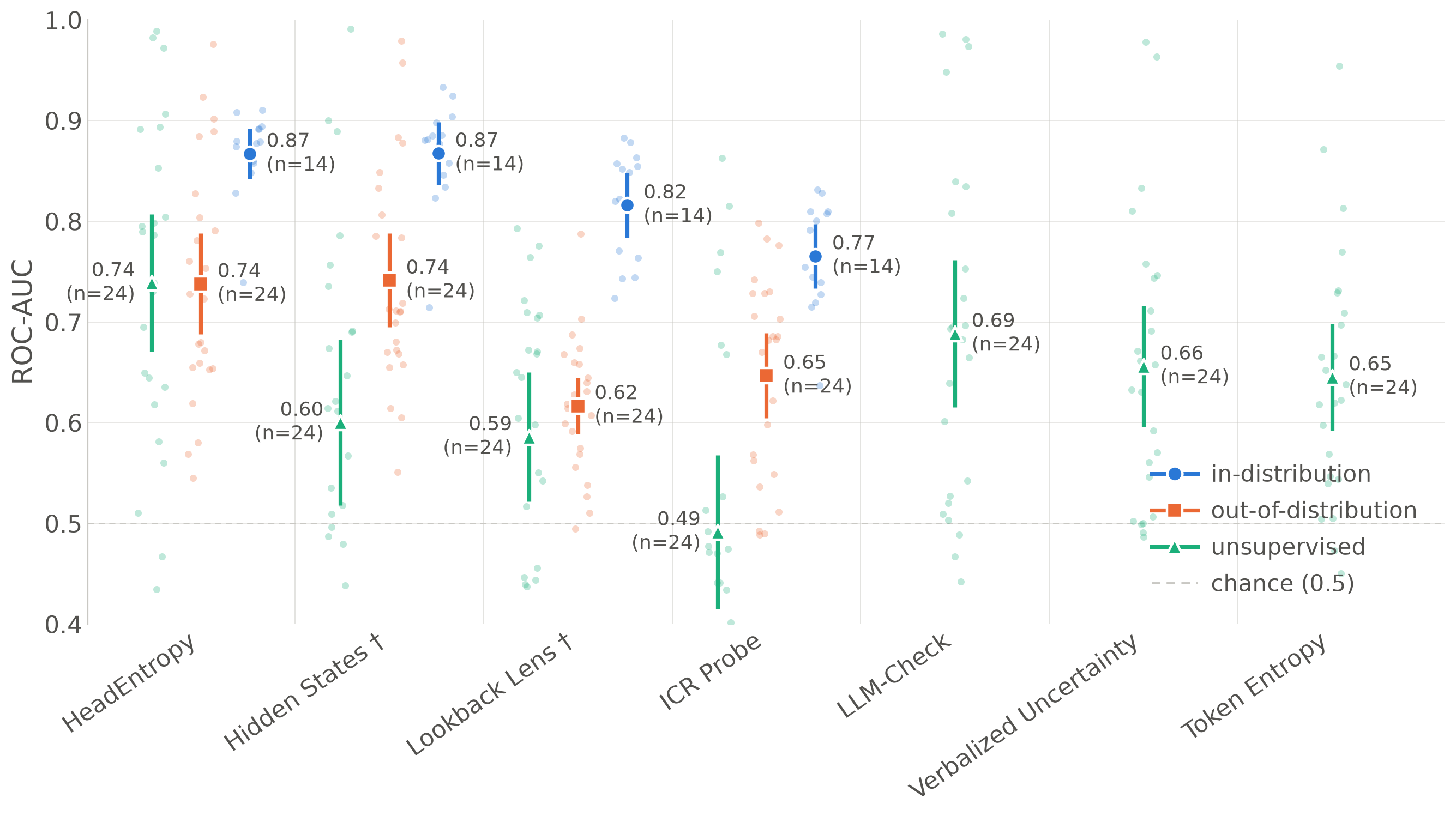}
    \caption{ROC-AUC by method and supervision regime. Each marker is the mean over (model, dataset) cells and the bar the 95\% interval of those cells, taken from the
    Student's $t$ quantile at $n-1$ degrees of freedom; the faint dots are the individual cells. In-distribution (trained probe) covers 14 cells and the other two regimes (Out-of-distribution/domain, training free) cover 24, since no probe is trained on MATH or FEVER. LLM-Check, verbalized uncertainty and token entropy exist only in the training-free regime and so carry a single marker.
    Rows marked $\dagger$ are plotted as $1-\mathrm{AUC}$: those features are
    anti-correlated with correctness in the training-free setting, and the flip was applied after seeing the results, whereas the orientation of \name{} is fixed a priori by Eq.~\ref{eq:training_free}. Figure~\ref{fig:paired-differences} gives the same comparison as within-cell paired differences.}
    \label{fig:auc-by-method-regime}
\end{figure}

Figure~\ref{fig:auc-by-method-regime} gives the absolute ROC-AUC behind the
paired differences of Figure~\ref{fig:paired-differences}: one marker per method
and supervision regime, with the individual (model, dataset) cells shown as faint
dots. In distribution, the \name{} probe (0.87) is level with the hidden-state
probe (0.87) and ahead of Lookback Lens (0.82) and ICR (0.77). Out of
distribution, \name{} keeps 0.74 while the hidden-state and Lookback Lens probes
fall to 0.62 and 0.65 and ICR to 0.49, close to chance. Without any probe,
\name{} again reaches 0.74, whereas the training-free means of the other features
lie between 0.59 and 0.69.

\section{Explaining Predictions with SHAP Values \label{app:computation_shaply}}

To attribute predictive performance to specific heads, we compute SHAP values~\citep{shapley1953value, lundberg2017unified} on the trained classifier:
\begin{equation}
\phi_k(f) = \frac{1}{m!} \sum_{\pi} \Big[ f(P_k^\pi \cup \{k\}) - f(P_k^\pi) \Big],
\end{equation}
where $\pi$ ranges over all feature orderings, $P_k^\pi$ is the set of features preceding $k$ in $\pi$, and $\phi_k(f)$ quantifies the contribution of head $k \in \{1,...,m\}$. 

\begin{figure}[H]
    \centering
    \includegraphics[width=0.5\linewidth]{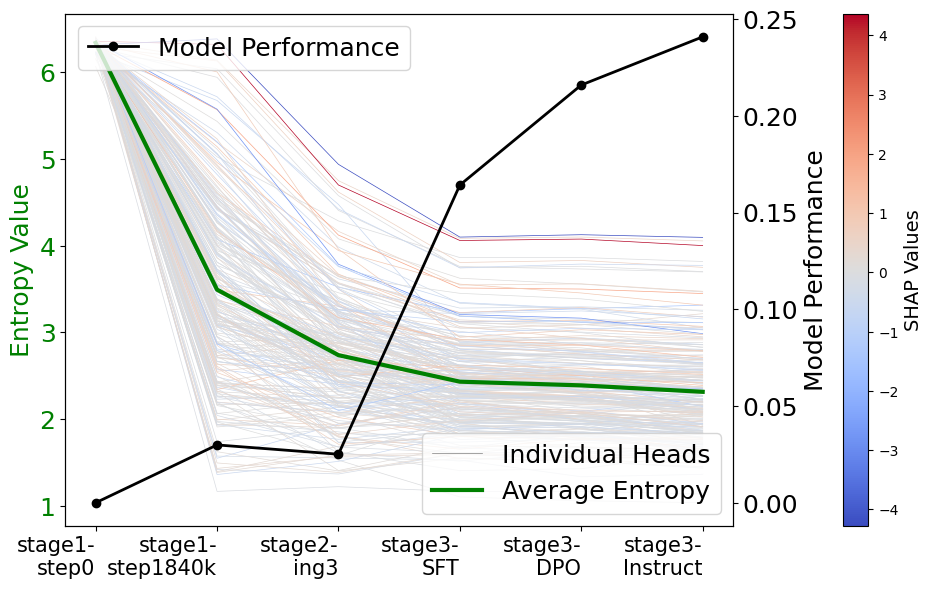}
    \caption{2-R\'enyi values (green) over training of OLMo-2-1B model. Each of the fine lines is a head's entropy values with color coding of the SHAP value (red means positive and blue means negative, more in Appendix \ref{app:computation_shaply}) of the final predictive logistic regression model. The green line represents the average entropy of the heads. Model Performance (black) measured in percentage of exact match examples for the validation set of TriviaQA.}
    \label{fig:entropy_training}
\end{figure}

Over the course of training we found that the heads with the highest entropy also carry the most predictive signal of the final model (Figure \ref{fig:entropy_training}). 

\section{Do predictive heads cluster in consistent depth ranges?}
\label{app:layer_consistency}

Our per-head SHAP analysis shows middle layers consistently carry higher-magnitude values (Table \ref{tab:shap-layer-depth}) with an average relative layer index of above-mean SHAP value layers at about 0.3-0.6. 

\begin{table}[H]
\centering
\caption{Average layer index of above-mean SHAP value layers, normalized by total number of layers.}
{%
\label{tab:shap-layer-depth}
\
\begin{tabular}{lccc}
\toprule
Model & TriviaQA & HotpotQA & MedMCQA \\
\midrule
Qwen3-1.7B    & 0.532 & 0.491 & 0.552 \\
Qwen3-8B      & 0.519 & 0.464 & 0.565 \\
Llama-3.1-8B  & 0.327 & 0.469 & 0.354 \\
Qwen3-32B     & 0.513 & --    & 0.489 \\
\bottomrule
\end{tabular}
}
\end{table}

\begin{figure}[H]
    \centering
    \includegraphics[width=\linewidth]{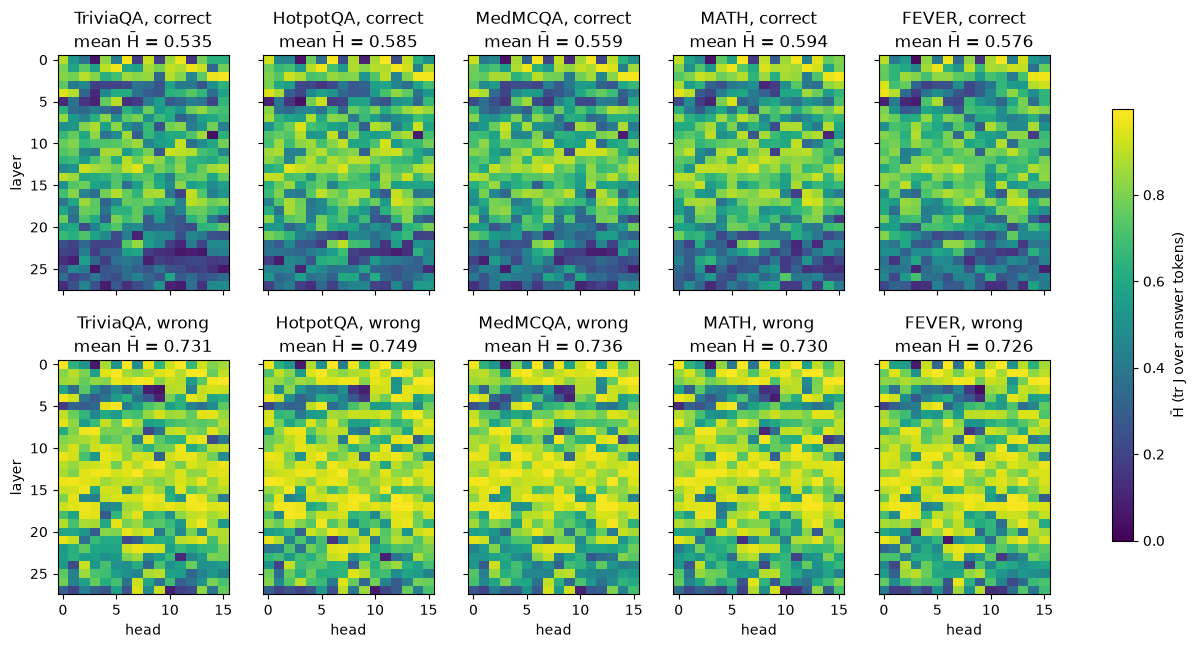}
    \caption{Trace of the Jacobian of the softmax per head for one correct and one incorrect example per dataset. For the wrong examples the heads are more yellow, especially the middle heads, indicating less settled attention patterns. }
    \label{fig:correct_wrong}
\end{figure}

Qualitative examples of each dataset for the Qwen3 1.7B Instruct model are shown in Figure \ref{fig:correct_wrong}.

\section{Contributions of Entropy across Layers}
\label{app:layer_ablation}
\begin{figure}[H]
    \centering
    \includegraphics[width=\linewidth]{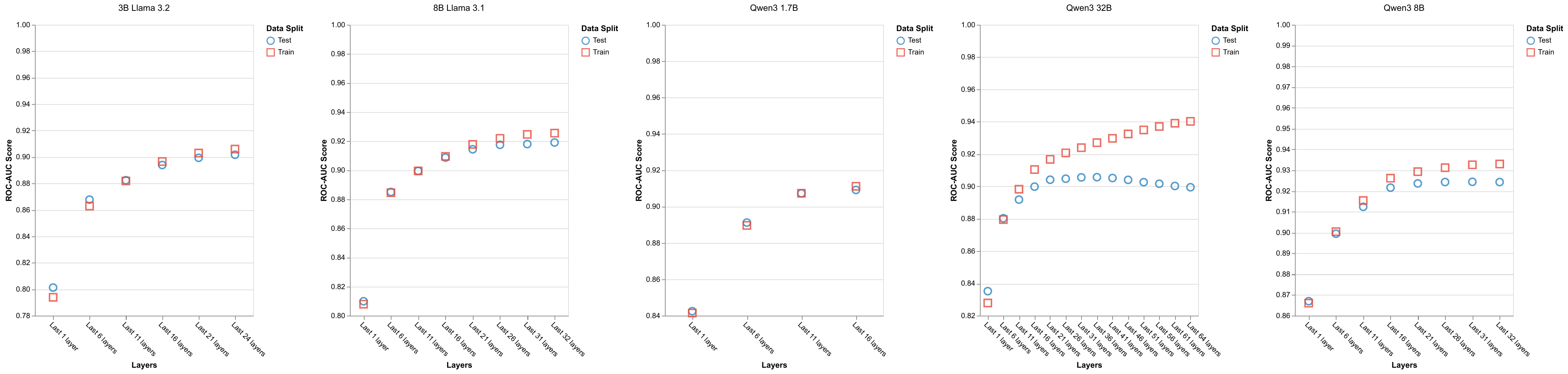}
    \caption{TriviaQA: Layer ablation, where layer 1 is the last layer and then adding more layers.}
    \label{fig:layers_triviaqa}
\end{figure}

\begin{figure}[H]
    \centering
    \includegraphics[width=\linewidth]{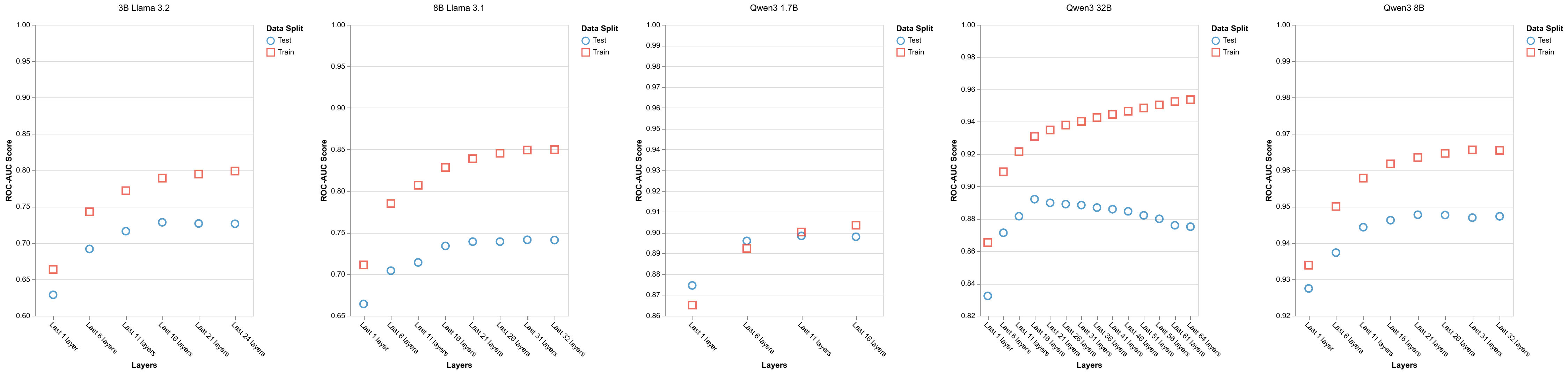}
    \caption{MedMCQA: Layer ablation, where layer 1 is the last layer and then adding more layers.}
    \label{fig:layers_medmcqa}
\end{figure}

\begin{figure}[H]
    \centering
    \includegraphics[width=\linewidth]{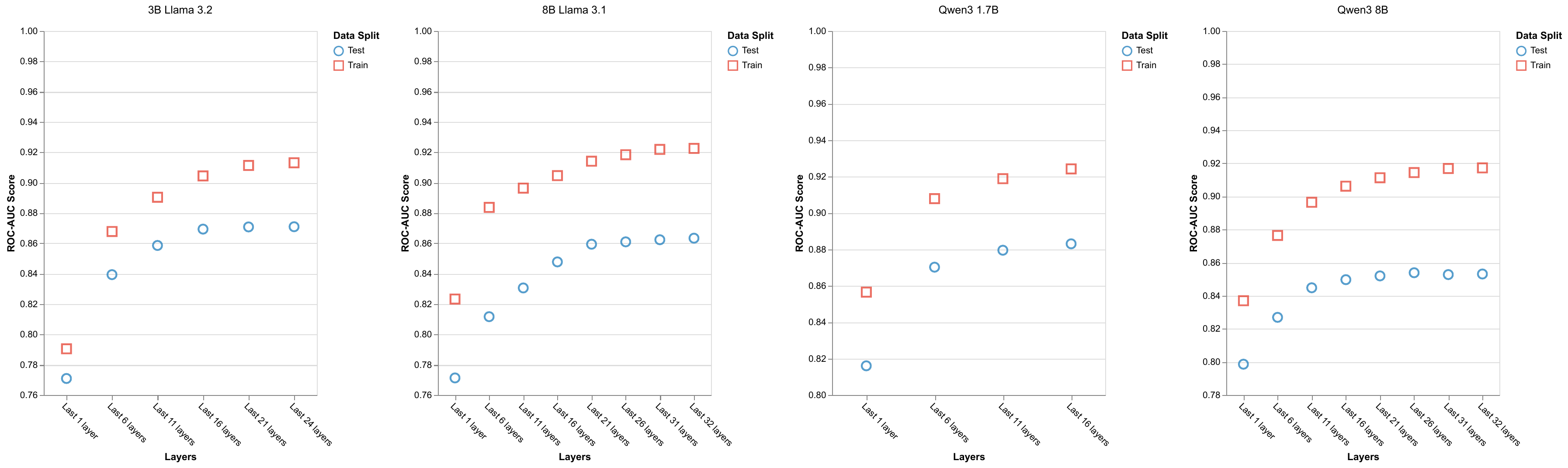}
    \caption{HotpotQA: Layer ablation, where layer 1 is the last layer and then adding more layers.}
    \label{fig:layers_hotpotqa}
\end{figure}

We conducted a layer ablation study, beginning with the final layer and progressively adding five layers at a time (Figures~\ref{fig:layers_triviaqa}, \ref{fig:layers_medmcqa} and \ref{fig:layers_hotpotqa} for TriviaQA, MedMCQA and HotpotQA). The results indicate that predictive value is not limited to the last layer alone: test AUROC rises over the first few added blocks and then plateaus well before all layers are included.

\section{Ablations}
This section collects the ablations referred to in Section~\ref{sec:results}: the regularization strength and training-set size of the probe, the aggregation of per-token entropies, the choice of probe, and the decoding temperature.
\subsection{Regularization Ablation}
\label{app:c_ablation}
\begin{figure}[H]
    \centering
    \includegraphics[width=0.48\linewidth]{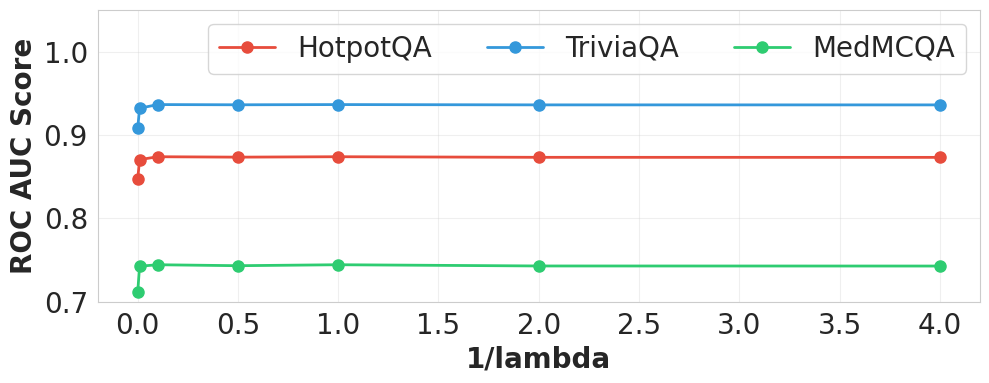}\hfill
    \includegraphics[width=0.48\linewidth]{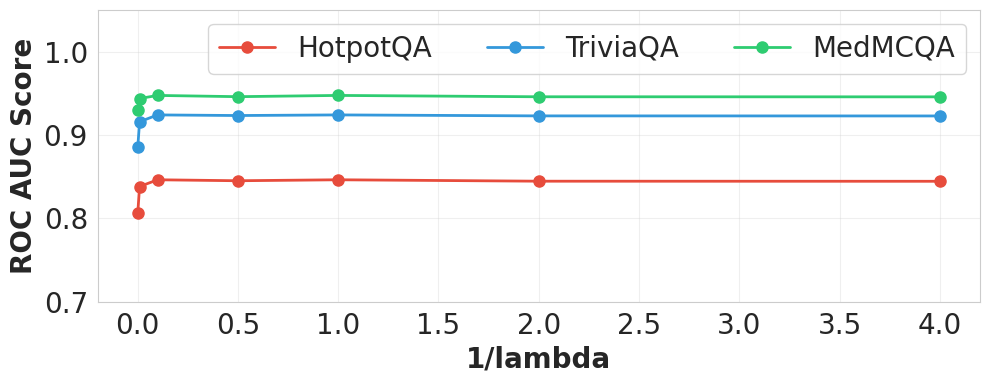}
    \caption{Performance vs. L1 regularization for Llama 3.1 8B (left) and Qwen3-8B (right), where for all datasets the performance converges at 0.1. Lower values correspond to more regularization strength.}
    \label{fig:c_ablation}
\end{figure}

We made the best effort to make the hyperparameter choice as standard as possible, as our main focus is methodological not empirical hyperparameter tuning. Please see the ablation in Figure \ref{fig:c_ablation}.
\subsection{Data Ablation}
\label{app:data_ablation}
\begin{figure}[H]
    \centering
    \includegraphics[width=0.48\linewidth]{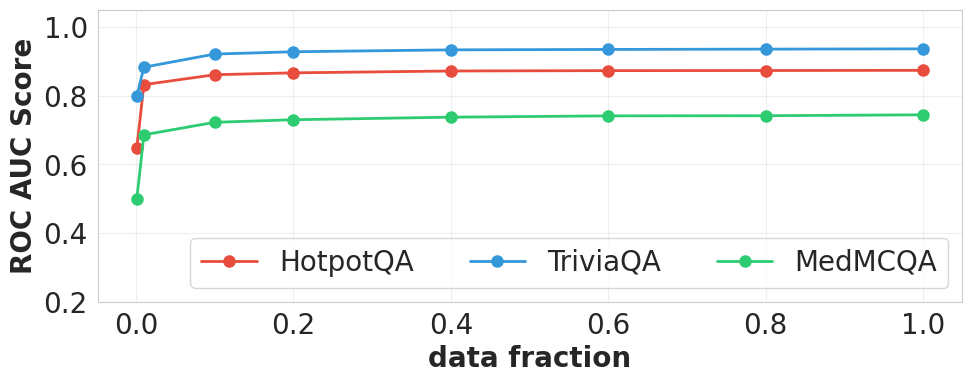}\hfill
    \includegraphics[width=0.48\linewidth]{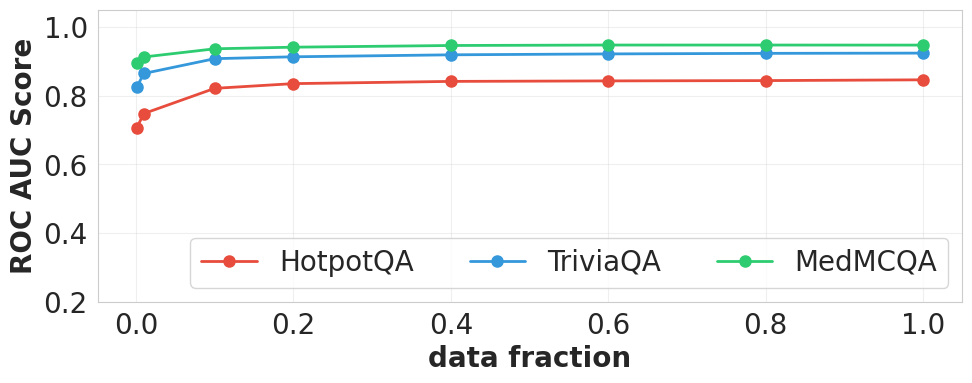}
    \caption{Performance vs. training set size for Llama 3.1 8B (left) and Qwen3-8B (right), where 1 data fraction corresponds to 50k examples. For all datasets the performance converges at around 0.2.}
    \label{fig:data_ablation}
\end{figure}

Please see the curves showing performance vs. training set size to demonstrate data efficiency Figure \ref{fig:data_ablation}. 

Experiments suggest \name{} maintains strong performance even with 20-30\% of training data.

\subsection{Aggregation across tokens}
\label{app:min_max_mean}
We aggregate per-token head entropies into a single per-head feature by averaging over the tokens of a semantic section (Eq.~\ref{eq:avg_head_entropy}). We compared the mean against the per-token minimum and maximum on Llama-3.1-8B and Qwen3-8B across TriviaQA, HotpotQA, and MedMCQA (Table~\ref{tab:min_max}). The mean is consistently the best or tied-best, followed closely by the maximum (within roughly $0.02$ AUROC), while the minimum is the weakest. We therefore use the mean throughout (Table \ref{tab:min_max}).
\begin{table}[H]
\centering
\caption{Empirically testing average, minimum and maximum aggregation over the rows of the attention matrix. Average showing consistently slightly higher AUROC, followed by maximum, ranging within 2 AUROC points.}
\label{tab:min_max}
{%
\begin{tabular}{lcccccc}
\toprule
& \multicolumn{3}{c}{Llama-3.1-8B} & \multicolumn{3}{c}{Qwen3-8B} \\
\cmidrule(lr){2-4} \cmidrule(lr){5-7}
Dataset & Avg & Max & Min & Avg & Max & Min \\
\midrule
HotpotQA & 0.87 & 0.86 & 0.85 & 0.85 & 0.83 & 0.82 \\
MedMCQA  & 0.74 & 0.73 & 0.73 & 0.95 & 0.87 & 0.73 \\
TriviaQA & 0.94 & 0.92 & 0.92 & 0.92 & 0.92 & 0.92 \\
\bottomrule
\end{tabular}
}
\end{table}

\subsection{Probe vs. MLP vs. XGBoost}
\label{app:xgboost_mlp}
We tested the AUROC on the validation split of each dataset for three probes trained on the per-head \name{} vector ($L \times H$ values, one per attention head, averaged over
the answer tokens) of the same dataset's train split: a logistic regression over all layers (LR, the probe
of the main text), a two-hidden-layer MLP ($1024$ units, ReLU, Adam, $20$ epochs) and gradient-boosted trees
(XGBoost, logistic objective, up to $100$ rounds) (Table \ref{tab:probes}).
\begin{table}[H]
\centering
\caption{Trained probes on the \name{} vector. Bold: best probe per model and dataset. The choice of
probe changes AUROC by at most a few points.}
\label{tab:probes}
{%
\begin{tabular}{llcccccc}
\toprule
Dataset & Probe & Qwen3 & Llama-3.2 & Qwen3 & Llama-3.1 & Gemma-4 & Qwen3 \\
        &       & 1.7B  & 3B        & 8B    & 8B        & 12B     & 32B \\
\midrule
TriviaQA & LR      & 0.894 & \textbf{0.924} & \textbf{0.892} & 0.879 & \textbf{0.908} & 0.877 \\
         & MLP     & \textbf{0.902} & 0.923 & 0.891 & \textbf{0.880} & 0.903 & \textbf{0.895} \\
         & XGBoost & 0.863 & 0.911 & 0.881 & 0.862 & 0.882 & 0.879 \\
\midrule
HotpotQA & LR      & 0.874 & 0.871 & \textbf{0.848} & 0.858 & \textbf{0.828} & -- \\
         & MLP     & \textbf{0.877} & \textbf{0.873} & 0.816 & \textbf{0.866} & 0.650 & -- \\
         & XGBoost & 0.866 & 0.855 & 0.839 & 0.836 & 0.816 & -- \\
\midrule
MedMCQA  & LR      & \textbf{0.891} & \textbf{0.722} & 0.911 & 0.739 & 0.879 & 0.860 \\
         & MLP     & 0.866 & 0.707 & \textbf{0.937} & \textbf{0.742} & \textbf{0.900} & 0.862 \\
         & XGBoost & 0.796 & 0.706 & 0.911 & 0.723 & 0.856 & \textbf{0.868} \\
\bottomrule
\end{tabular}
}
\end{table}

\subsection{Robustness to decoding temperature}
\label{app:temperature}

Table~\ref{tab:temperature-ablation} reports AUROC for a \name{}
classifier trained at $T{<}0.0001$ and evaluated on generations
sampled at $T{=}1$. 

\begin{table}[H]
\centering
\caption{Performance under different temperature at inference time remains comparable. For the MedMCQA dataset the overall performance drops below random (0.25), likely responsible for the performance drop in AUROC of \name{}.}
\label{tab:temperature-ablation}
{%
\begin{tabular}{llcccc}
\toprule
& & \multicolumn{2}{c}{AUROC of \name{}} & \multicolumn{2}{c}{Overall Model Accuracy} \\
\cmidrule(lr){3-4} \cmidrule(lr){5-6}
Model & Dataset & $T < 0.0001$ & $T = 1$ & $T < 0.0001$ & $T = 1$ \\
\midrule
\multirow{3}{*}{Llama-3.1-8B}
 & TriviaQA  & 0.93 & \bfseries 0.94 & 0.45 & 0.40 \\
 & MedMCQA   & 0.74 & 0.54  & 0.50 & 0.18 \\
 & HotpotQA  & 0.88 & \bfseries 0.89 & 0.38 & 0.36 \\
\midrule
\multirow{3}{*}{Qwen3-8B}
 & TriviaQA  & 0.92 & \bfseries 0.91 & 0.58 & 0.53 \\
 & MedMCQA   & 0.94 & 0.70 & 0.39 & 0.12 \\
 & HotpotQA  & 0.80 &  \bfseries 0.85 & 0.58 & 0.56 \\
\bottomrule
\end{tabular}
}
\end{table}


\FloatBarrier

\section{Pseudocode}
\label{app:head_entropy_algorithm}
Algorithm~\ref{alg:head_entropy} summarizes the full \name{} pipeline: for each example, we compute the 2-R\'enyi entropy of every attention head at every token, convert it to the trace of the softmax Jacobian (Eq.~\ref{eq:head_entropy}), average over the tokens of a semantic section, stack into a per-example feature vector, and train an $\ell_1$-regularized logistic regression on the binary correctness label.

\begin{algorithm}[H]
\caption{\name{}}
\label{alg:head_entropy}
\begin{algorithmic}
\REQUIRE Dataset $\{(x_i, y_i)\}_{i=1}^N$; transformer with $m$ heads; sequence length $T$ of $x_i$
\FOR{each $(x_i, y_i)$}
    \STATE Obtain $\hat{y}_i$ and attention tensor $A_i \in \mathbb{R}^{m \times T \times T}$ (Eq.~\ref{eq:attention_matrix})
    \FOR{each head $k$ and token $t$}
        \STATE Compute 2-R\'enyi entropy $H_{k,i}^{(t)} = -\log \sum_{j} \left(A_{k,i}^{(t,j)}\right)^2$
        \STATE Compute $\operatorname{trace}(J)_{k,i}^{(t)} = 1 - \exp(-H_{k,i}^{(t)})$ \hfill (Eq.~\ref{eq:head_entropy})
    \ENDFOR
    \STATE Average over section tokens to get $\bar{H}_{k,i} \in \mathbb{R}$ \hfill (Eq.~\ref{eq:avg_head_entropy})
    \STATE Stack into feature vector $\bar{H}_i = [\bar{H}_{1,i},\dots,\bar{H}_{m,i}]^\top \in \mathbb{R}^{m}$
\ENDFOR
\STATE Stack $\{\bar{H}_i\}$ into feature matrix $\bar{H} \in \mathbb{R}^{N \times m}$
\STATE Define labels $z_i = \mathbb{I}[\hat{y}_i = y_i]$, where $z \in \mathbb{R}^{N}$
\STATE Take the mean of $\bar{H}$ across $m$ $\bar{H}_{\text{mean}}$
\RETURN $\bar{H}_{\text{mean}}$, $z$. 
\end{algorithmic}
\end{algorithm}

\section{Derivation of the Softmax Jacobian}
\label{app:softmax_jacobian}

We derive the Jacobian of the softmax function. Let:
\begin{equation}
    p_i := \mathrm{softmax}(x)_i = \frac{\exp(x_i)}{\sum_{j=1}^K \exp(x_j)}
\end{equation}

To compute $\frac{\partial p_i}{\partial x_j}$, we consider two cases.

\paragraph{Case 1: $i = j$.} We apply the quotient rule:
\begin{align}
    \frac{\partial p_i}{\partial x_i} &= \frac{\exp(x_i) \cdot \sum_k \exp(x_k) - \exp(x_i) \cdot \exp(x_i)}{\left(\sum_k \exp(x_k)\right)^2} \\
    &= \frac{\exp(x_i)}{\sum_k \exp(x_k)} - \frac{\exp(x_i)^2}{\left(\sum_k \exp(x_k)\right)^2} \\
    &= p_i - p_i^2 = p_i(1 - p_i)
\end{align}

\paragraph{Case 2: $i \neq j$.} The numerator $\exp(x_i)$ does not depend on $x_j$, so:
\begin{align}
    \frac{\partial p_i}{\partial x_j} &= \frac{0 - \exp(x_i) \cdot \exp(x_j)}{\left(\sum_k \exp(x_k)\right)^2} = -p_i p_j
\end{align}

\paragraph{Combined.} We can write the Jacobian compactly as:
\begin{equation}
    \mathcal{J}_{\mathrm{softmax}} = \mathrm{diag}(p) - pp^\top
\end{equation}

We can now see that the trace of the Jacobian is monotonically related to the 2-R\'enyi entropy.

Recall that 2-R\'enyi entropy is $H_2(p) := -\log \sum\limits_{i}p_i^2$. Using $J = \mathrm{diag}(p) - pp^T$:

\begin{align*}
Tr(J)&=Tr(\mathrm{diag}(p)) - Tr( pp^T) \\
&= \sum\limits_i p_i - ||p||^2\\
&= 1 - ||p||^2\\
&= 1 - \exp(-H_2(p))
\end{align*}

\FloatBarrier

\section{LLM contributions}
In this work, we used LLMs to help improve the readability of the paper and create figures. 

\end{document}